Article

# A Data-Driven Two-Phase Multi-Split Causal Ensemble Model for Time Series

Zhipeng Ma [1,2], Marco Kemmerling [2,*], Daniel Buschmann [3], Chrismarie Enslin [2], Daniel Lütticke [2] and Robert H. Schmitt [2,3,4]

1. SDU Center for Energy Informatics, The Maersk Mc-Kinney Moller Institute, University of Southern Denmark, Campusvej 55, 5230 Odense, Denmark; zhma@mmmi.sdu.dk
2. Information Management in Mechanical Engineering, RWTH Aachen University, Dennewartstraße 27, 52068 Aachen, Germany
3. Laboratory for Machine Tools and Production Engineering WZL, RWTH Aachen University, Campus-Boulevard 30, 52074 Aachen, Germany
4. Fraunhofer Institute for Production Technology (IPT), Steinbachstraße 17, 52074 Aachen, Germany

Correspondence: marco.kemmerling@ima.rwth-aachen.de; Tel.: +49-241-80-911-83

**Abstract:** Causal inference is a fundamental research topic for discovering the cause–effect relationships in many disciplines. Inferring causality means identifying asymmetric relations between two variables. In real-world systems, e.g., finance, healthcare, and industrial processes, time series data from sensors and other data sources offer an especially good basis to infer causal relationships. Therefore, many different time series causal inference algorithms have been proposed in recent years. However, not all algorithms are equally well-suited for a given dataset. For instance, some approaches may only be able to identify linear relationships, while others are applicable for non-linearities. Algorithms further vary in their sensitivity to noise and their ability to infer causal information from coupled vs. non-coupled time series. As a consequence, different algorithms often generate different causal relationships for the same input. In order to achieve a more robust causal inference result, this publication proposes a novel data-driven two-phase multi-split causal ensemble model to combine the strengths of different causality base algorithms. In comparison to existing approaches, the proposed ensemble method reduces the influence of noise through a data partitioning scheme in a first phase. To achieve this, the data are initially divided into several partitions and the base causal inference algorithms are applied to each partition. Subsequently, Gaussian mixture models are used to identify the causal relationships derived from the different partitions that are likely to be valid. In the second phase, the identified relationships from each base algorithm are then merged based on three combination rules. The proposed ensemble approach is evaluated using multiple metrics, among them a newly developed evaluation index for causal ensemble approaches. We perform experiments using three synthetic datasets with different volumes and complexity, which have been specifically designed to test causality detection methods under different circumstances while knowing the ground truth causal relationships. In these experiments, our causality ensemble outperforms each of its base algorithms. In practical applications, the use of the proposed method could hence lead to more robust and reliable causality results.

## 1. Introduction

To understand, modify, and potentially improve complex real-world systems, such as production processes, ecosystems, and nervous systems, it is essential to study the internal structure of relationships between different components of a system and understand why the system exhibits certain behaviors. Research in the direction of causal inference thematically addresses such problems. Causality refers to the relationship between causes and effects [1,2], where the cause is responsible for the effect and the effect is dependent on the cause. Such one-directional influence represents an asymmetric relationship. Since process data in real-world systems are mostly time series [3], we focus on causal modeling in a time series setting. Here, the asymmetric nature of causality becomes especially apparent: the past can influence the future, but the future cannot influence the past.

Some causal inference methods have been proposed to explore cause–effect relationships between different variables of a system based on time series datasets. The Granger causality test (GC) [4] is the first causal inference approach for time series, which is based on statistical hypothesis testing. Transfer entropy (TE) [5] is a non-parametric statistic measurement based on information theory to identify non-linear causal relationships. Sugihara et al. [6] applied Takens' embedding theorem to propose convergent cross mapping (CCM) for detecting causality in complex dynamic systems. In these approaches based on the systems' predictability, time series data with two or more variables are taken as input, and the forecast on the cause–effect relationship among these variables is produced.

However, in many cases distinct causal inference approaches produce different causality relationships based on the same inputs [7,8]. When performing causal inference on a given dataset, it is, hence, not clear which causal inference algorithm should be chosen, since clearly no single algorithm is completely reliable. This is the main challenge we address in this publication. Differences in causal results can arise due to a variety of reasons. For instance, the Granger causality test works well at predicting linear relationships but does not perform well in non-linear systems. CCM can be applied to investigate strongly coupled complex systems, but it misses simple linear causal relationships in many examples. Real-world datasets often feature diverse underlying relationships. In such datasets, not only linear and weakly coupled relationships, but also non-linear and strongly interacting feature pairs are integrated. Recent studies aim to address this complexity by extending the classic causal inference algorithms with machine/deep learning frameworks. Tank et al. [9] proposed the neural Granger causality algorithm by applying structured multilayer perceptrons (MLPs) or recurrent neural networks (RNNs) combined with sparsity-inducing penalties on the weights, which is a powerful non-linear extension of the GC. Clark et al. [10] combine the CCM with dewdrop regression to build a novel causality test scheme that leverages spatial replication, resulting in good performance in the application of short and simple time series.

Next, the performance of many causal reasoning approaches is limited when processing complex large-scale datasets because of the observational error and process noise [6]. To improve the performance of causality detection, Peng et al. [11] developed an interpretable deep learning architecture for the GC estimation that integrates long short-term memory (LSTM), autoencoder (AE), and a convolutional neural network (CNN). Furthermore, many algorithms, such as GC and graphical causal models [12], produce qualitative causal results, but quantitative ones can provide more detailed and clearer causal information [8]. Porta et al. [13] quantified the causality strength among multiple components in network physiology and checked the reliability of the GC pipeline using hypothesis testing. Zaremba et al. [14] quantified causality relationships through Gaussian process models. In such an analysis, a nested model is formulated for deducing causality in both mean and covariance through automatic relevance determination (ARD) construction of the kernel.

To address the issue of differing causality results from different methods, some approaches try to combine the strengths of multiple individual methods by applying ensemble learning techniques. Ensemble learning [15] is a typical model combination method in statistics and machine learning that has proven effective at providing equal or better results than any individual algorithm. In the domain of causal reasoning specifically, it is known that more reasonable causal relationships can be extracted through the combination process of ensemble learning [16], which addresses the inconsistency of different causal inference algorithms when applied to the same input. Such causality ensembles have previously been investigated by Li et al. [17], who utilized a bagging mechanism with a new weighting criterion to fuse different Bayesian network (BN) structures. This framework has higher accuracy and a more powerful generalization ability than a single

BN-learner. Athey et al. [18] proposed a weighted averaging model with weights determined by stacked regression based on cross validation, to ensemble three base causal algorithms. Guo et al. [19] introduced a flexible two-phase ensemble framework by utilizing data partitioning and majority voting. The mentioned causal ensemble models are all based on the conventional ensemble strategies, such as majority voting and bagging, and most of them are one-phase frameworks, of which the accuracy is not high in some complex experiments. Hence, a more comprehensive ensemble framework can be developed to take full advantage of distinct base learners. This study developed a novel two-phase ensemble framework, embedding softer classification strategies than majority voting and bagging compared with the existing models. The proposed model adopted the split-and-ensemble processing aimed at taking advantage of the causal information of the base algorithms and reducing the noise effect in the large-scale datasets, which improves the accuracy and stability of the causality detection.

To improve the limitations mentioned above, this paper proposes a stable and reliable data-driven multi-split two-phase causal ensemble model for time series. It combines state-of-the-art causal inference methods for time series, involving the GC [4], normalized transfer entropy (NTE) [20], PCMCI+ [21], and CCM [6]. To make full use of the valuable information from the datasets and the results of multiple base learners, we develop a split-and-ensemble framework, where the dataset is split into several partitions and combined through a two-phase ensemble scheme. The split-and-ensemble process reduces the influence of noise in the data, and the two-phase ensemble steps take advantage of all base causal inference algorithms to cope with the inconsistency of their causal results under the same input. Since real-world datasets typically lack a ground truth label for causal relationships, we perform the evaluation of our approach on three synthetic datasets with different volumes and complexity.

Reflecting on the outlined scope of this publication, our key contributions pertain to the following aspects:

- To address the issue of distinct causal inference methods producing different results, we introduce a multi-split two-phase causal ensemble model for time series data. This model
  - Reduces the influence of noise through a data partitioning scheme;
  - Combines the advantages of distinct causal inference algorithms to arrive at stronger results than each of them individually;
  - Is flexible, which means that the number of base learners can be adjusted and the chosen causal inference methods can be replaced by any other causality discovery models.
- To assess the credibility of causality results from ensembles such as our approach, we propose a novel evaluation index.

The structure of the paper is organized as follows. In Section 2, the individual causal inference methods which are used in our ensemble approach are introduced. In Section 3, the proposed causality ensemble model and the components are elaborated on and demonstrated in detail. Section 4 presents the evaluation index towards the proposed model. Section 5 explains the experiments on the synthetic datasets, as well as corresponding results and discussions. Finally, Section 6 concludes the research and presents recommendations for future work.

## 2. Causal Inference Algorithms for Time Series

In this section, the principle of four causal inference algorithms for time series data, namely GC (Section 2.1), NTE (Section 2.2), PCMCI+ (Section 2.3), and CCM (Section 2.4), are introduced in detail. We have selected these four models as the base learners of the proposed ensemble model because of their different properties. These different properties are rooted in the distinct internal mechanisms of the algorithms that each address certain failings of causal models. Their detailed advantages and limitations, as well as how these are addressed in the ensemble, are elaborated in Section 2.1 to Section 2.4. Note that the selected approaches only serve as examples for our ensemble approach. In principle, any causal inference method can be employed in the ensemble.

*2.1. Granger Causality Test*

The GC has been widely used for causal inference in time series analysis since its introduction in 1969 [22]. It is a statistical method that can identify linear causal relationships in a proposed causal model.

Clive Granger defined causality by a linear prediction model for stochastic processes [4]. The general idea behind it is that a time series $Y$ causing $X$ contains unique information about $X$, helping explain its future trend. This method mainly focuses on linear relations and processes instantaneous effects. The time series $X$ *Granger-causes* another time series $Y$, if, through statistical hypothesis tests (e.g., $t$-test [23] or F-test [24]) on lagged values $X$ and $Y$, statistically significant information is provided about future values of $Y$. It signifies that the prediction of $Y$ improves by incorporating $X$'s past values $X_{past(t)}$ into its own history $Y_{past(t)}$. Equation (1) is the mathematical expression of the GC and Equation (2) [22,25] is the hypothesis to be tested.

$$GrangerCausality_{X \to Y} \iff Y_t \not\perp\!\!\!\perp X_{past(t)} \mid Y_{past(t)} \tag{1}$$

where $\iff$ refers to "if and only if", $\not\perp\!\!\!\perp$ represents "not independent", and $\mid$ is the condition symbol.

$$\mathbb{P}[Y_{t+n} \in A \mid IN(t)] \neq \mathbb{P}[Y_{t+n} \in A \mid IN_Y(t)] \tag{2}$$

where $n \in \mathbb{Z}^+$, $\mathbb{P}$ refers to the probability, $A$ represents an arbitrary non-empty set, $IN(t)$ denotes the information available at time $t$ in the entire universe including $X$ and $Y$, and $IN_Y(t)$ is the information only from $Y$ at time $t$.

Vector autoregressive model (VAR) [26] is typically fitted for the Granger causality analysis.

$$\hat{Y}_{t+1} = \sum_{i=0}^{n-1} \alpha_i Y_{t-i} + \epsilon_{Y,t+1} \tag{3}$$

$$\hat{Y}_{t+1} = \sum_{i=0}^{n-1} a_i X_{t-i} + \sum_{i=0}^{n-1} b_i Y_{t-i} + \epsilon_{Y|X,t+1} \tag{4}$$

where $\alpha_i$, $a_i$, and $b_i$ are coefficients of models, $\epsilon_Y$ and $\epsilon_{Y|X}$ refer to noise terms. If the variance $var(\epsilon_{Y|X})$ is significantly smaller than $var(\epsilon_Y)$, $Y$ is said to be Granger-caused by $X$.

In application, the VAR model is utilized for linear regression. After prediction based on Equations (3) and (4), a hypothesis test is performed to determine if the lagged values $X$ in the causal pair $(X, Y)$ significantly influence $Y$. The hypothesis is based on Equation (2), and a $t$-test or F-test is conducted to determine the statistical significance of the results based on a preset significance level. When the hypothesis is accepted, it is concluded that the regression function in Equation (3) performs better than that in Equation (4), so $Y$ is caused by $X$. The absolute correlation coefficient between the estimated values $\hat{Y}$ and $Y$ is computed as the causal strength.

In this study, we employed the Python library statsmodels to perform the Granger causality test [27].

To conclude, while the GC is a powerful tool for detecting linear causal relationships in time series analysis, it may not be able to identify non-linear causality due to its reliance on linear VAR models. To overcome this limitation, the proposed ensemble model incorporates three additional algorithms that are better suited for detecting non-linear causal relationships.

*2.2. Normalized Transfer Entropy*

The Transfer entropy (TE) [5] is capable of detecting non-linear causal relationships, which is a good supplement of the GC. The NTE [20] normalizes the causality strength to the range 0–1 prepared for the integration steps in the proposed ensemble model.

Information theory is a prominent research domain to analyzing the information flow between two processes in time order. The TE [5] is a non-parametric statistic measurement and a basic method for inferring non-linear causality connections, which is the conditional mutual information (CMI) given the past values of the influenced variable. If the amount of information is measured using Shannon's entropy, the TE from a time series $X$ to another one $Y$ can be defined as follows

$$\begin{aligned} TE_{X \to Y} &= I(Y_t; X_{t-1:t-L} | Y_{t-1:t-L}) \\ &= H(Y_t | Y_{t-1:t-L}) - H(Y_t | Y_{t-1:t-L}, X_{t-1:t-L}) \end{aligned} \quad (5)$$

where $TE_{X \to Y}$ is the TE from $X$ to $Y$. $I(x)$ represents CMI, and $H(x|y)$ represents the conditional Shannon entropy given in Equation (6).

$$H(X|Y) = -\sum_{x,y} p(x,y) \log p(x|y) \quad (6)$$

where $p(x,y)$ is the joint probability density function and $p(x|y)$ denotes the conditional probability density.

When $TE_{X \to Y} > 0$, $X$ is seen to be the cause of $Y$ and the causal strength becomes stronger with the increase in transfer entropy. However, since it is expensive to compute high-dimensional probability density, approximation approaches should be developed for an efficient application. To make the TE a faster and more efficient causality assessment tool, Ikegwu et al. [28] investigated a new estimation implementation with Kraskov's estimator and multiple processes. They accomplished this by parallelizing queries on k-dimensional trees, which are the binary search trees for organizing k-dimensional data points. This approach significantly reduces the wall time. A. Kraskov [29] estimated entropy based on kNN:

$$\hat{H}(X) = -\frac{1}{n}\sum_{i=1}^{n} \psi(n_x(i)) - \frac{1}{k} + \psi(n) + \ln(c_{d_x}) + \frac{d_x}{n}\sum_{i=1}^{n} \ln(\epsilon(i)) \quad (7)$$

where $\hat{H}(X)$ represents the estimated entropy, $n$ is the number of data instances, $k$ is the number of nearest neighbors, $d_x$ refers to the dimension of $x$, $c_{d_x}$ represents the volume of the $d_x$-dimensional unit ball, and $\frac{\epsilon(i)}{2}$ denotes the distance between the $i$-th data point and its $k$-th neighbor. $n_x(i)$ refers to the number of data instances inside the unit ball, which involves all the points $x_j$, such that $\|x_i - x_j\| \le \frac{\epsilon(i)}{2}$, and $\psi(x)$ comprises the digamma function where

$$\psi(x) = \frac{\Gamma'(x)}{\Gamma(x)} \quad (8)$$

where $\Gamma(x)$ denotes the gamma function.

Then, the mutual information is estimated based on $\hat{H}(X)$, $\hat{H}(Y)$, and $\hat{H}(X,Y)$.

$$\begin{aligned} \hat{I}(X,Y) &= \hat{H}(X) + \hat{H}(Y) - \hat{H}(X,Y) \\ &= \psi(k) - \frac{1}{k} + \frac{1}{n}\sum_{i=1}^{n} [\psi(n_x(i)) + \psi(n_y(i))] + \psi(n) \end{aligned} \quad (9)$$

where $\hat{I}(X,Y)$ is the estimated mutual information, and $n_y(i)$ refers to the number of all data points $y_j$, such that $\|y_i - y_j\| \le \frac{\epsilon_y(i)}{2}$.

The range of the TE (both analytical and estimated form) is $[0, +\infty)$, which is inconsistent with the causal strength from the other three learners (GC [4], PCMCI+ [21], and CCM [6]), all of which have ranges of $[0, 1]$. Therefore, the NTE [20] is proposed to trans-form the TE values into a form with a range of $[0, 1]$. The NTE value from variable $X^i$ to $X^j$ is

$$NTE_{X^i \to X^j} = \frac{TE_{X^i \to X^j} - TE_{X^i \to X^j}^{Shuffled}}{H(X_t^j | X_{t-1:t-\tau_{max}}^j)} \quad (10)$$

where $TE_{X^i \to X^j}$ refers to the TE value from $X^i$ to $X^j$. $TE_{X^i \to X^j}^{Shuffled}$ is the shuffled TE, where the values in $X^i$ are drawn in random order. $H(X_t^j | X_{t-1:t-\tau_{max}}^j)$ represents the conditional entropy of $X^j$ at time $t$ given its past values. The subtraction of the shuffled TE overcomes the bias in the TE calculation due to finite data size and limited time delay [30]. We use the Python library PyIF [28] for the TE implementation.

As data complexity and volume increases, the NTE may fail to identify some causal results due to the uncertainty in the estimation process. The following sections describe PCMCI+ and CCM, which are especially suitable for complex and large-scale systems and aim to balance out the weaknesses of NTE.

*2.3. PCMCI+*

PCMCI [31] is a state-of-the-art causality detection technique for processing large-scale complex datasets, which provides convinced causal results in extremely complex systems. PCMCI+ [21] extents PCMCI by detecting contemporaneous (also called instantaneous) links, resulting in more reliable causal reasoning.

PCMCI takes advantage of graphical models to estimate causality structures from time series data. This occurs in a time-dependent system $\mathbf{X_t} = (X_t^1, \ldots, X_t^N)$ with

$$X_t^i = f_i(P(X_t^i), \eta_t^i) \quad (11)$$

where $f_i$ represents the dependency function, $\eta_t^i$ is the random noise, and $P(X_t^i)$ denotes the causal parents of variable $X_t^i$ among the past of all $N$ features. If $X_{t-\tau}^j \in P(X_t^i)$, $X_{t-\tau}^j$ is said to be the cause of $X_t^i$ with $\tau$ delay.

PCMCI involves two steps. The PC algorithm [32] (named after its inventors) is the first stage, which is an iterative Markov discovery mechanism for condition selection. After the conditional independence (CI) tests, an estimate $\widehat{P}(X_t^i)$, a superset of the parents $P(X_t^i)$, is obtained for all variables in $\mathbf{X_t}$. In the second stage, the momentary conditional independence test (MCI) [33] is applied. The estimated parents in step one are embedded into the conditions of MCI. All variable pairs $(X_t^i, X_{t-\tau}^j)$ with $i, j \in \{1, \ldots, N\}$ and time delays $\tau \in \{1, \ldots, \tau_{max}\}$ being tested, the causal link $X_{t-\tau}^j \to X_t^i$ is established if, and only if,

$$X_{t-\tau}^j \not\perp\!\!\!\perp X_t^i \mid \widehat{P}(X_t^i) \setminus \{X_{t-\tau}^j\} \widehat{P}_{px}(X_{t-\tau}^j) \quad (12)$$

where "$\not\perp\!\!\!\perp$" represents "not independent", "$|$" is the condition symbol, and $\widehat{P}(X_t^i) \setminus \{X_{t-\tau}^j\}$ denotes that the estimate $\widehat{P}(X_t^i)$ excluding the values $X_{t-\tau}^j$.

PCMCI+ [21] is the extension of PCMCI, including the discovery of contemporaneous links. In applications, both lagged and contemporaneous causal links are significant, but the original PCMCI paradigm only analyzes the time-lag causal effects such as $X_{t-\tau}^i \to X_t^j (\tau > 0)$, resulting in missing causal information. In this algorithm, certain orientation phases are utilized to orient the instantaneous adjacencies $X_t^i - X_t^j$ so that the performance of the PC stage is enhanced. Firstly, the collider orientation phase orients the triple $X_{t-\tau}^k - X_t^i - X_t^j$. If $X_{t-\tau}^k$ and $X_t^j$ are independent after the CI test, the direction should be $X_{t-\tau}^k \to X_t^i \leftarrow X_t^j$ (for $\tau > 0$, we always have $X_{t-\tau}^i \to X_t^j$, because only the past can forecast the future). Secondly, the rule orientation phase introduces several rules to determine the arrows in the triple $X_t^k - X_t^i - X_t^j$. The experiments [21,34] demonstrate that it can be extremely useful in many real-world application scenarios in which time strong autocorrelation is present and resolutions are too coarse to resolve time delays.

We use the PCMCI+ implementation of the Python library tigramite [21] in this study.

Although the PCMCI+ algorithm is a powerful tool for causal inference in time series data, it may fail to detect some causal relationships between coupled time series. To mitigate this issue, CCM is incorporated as a base learner in our model, which is well-suited for detecting causal relationships in non-linear and non-stationary systems.

*2.4. Convergent Cross Mapping*

Convergent Cross Mapping (CCM) [6] is a novel causal reasoning method based on the theory of non-linear state space reconstruction. This method works well in coupled time-series systems where the variables have interacting effects. Because GC, NTE, and PCMCI+ all have the limitations in analyzing coupled time series, which can result in missed causal relationships, CCM is a necessary component in our ensemble model.

In a dynamic system, a manifold [35] represents a topological space that locally resembles the Euclidean space near each data point of this system. The attractor manifold is the approximation of the dynamic system based on the real data, whereas the shadow manifold is a lower-dimensional approximation. For instance, considering a 3-dimensional system $S = [X, Y, Z]$, the attractor manifold is constructed based on the information from $X$, $Y$, and $Z$. The shadow manifold is estimated based on only $X$ or $Y$.

Takens' embedding theorem [36] states that an attractor manifold of a dynamical system can be reconstructed from a single observation variable of the system by a generic function. When the two time series $X$ and $Y$ belong to the same dynamics system, the corresponding shadow manifolds $M_X$ and $M_Y$ are diffeomorphic (have a one-to-one mapping) to the true manifold $M$. Data points close to each other on the manifold $M_X$ are also in a close distance on $M_Y$. Thus, the current state of the process $Y$ can be forecast based on $M_X$. $\hat{Y}|M_X$ is defined as the estimation of $Y$ under the condition of $M_X$. After several iterations of the cross mapping, a causal link $X \to Y$ can be declared if correlation coefficients of $\hat{Y}$ and $Y$ are convergent.

The procedure of the CCM is based on a k-nearest-neighbor algorithm (kNN) involving exponentially weighted distances from nearby points on a reconstructed manifold to execute kernel density estimations (KDE). Considering the time series $\mathbf{X} = (X^1, X^2, \ldots, X^N)$, for any two time series $(X^i, X^j)$ with length $L$ in $\mathbf{X}$, the shadow manifold is set by forming the lagged-coordinate vectors. For instance, the shadow manifold $\mathbf{M_{X_i}}$ is constructed in line with the vector $x_i(t) = [X_i(t), X_i(t-\tau), X_i(t-2\tau), \ldots, X_i(t-(E-1)\tau)]$ for $t = 1 + (E-1)\tau$ to $L$. Then, $\hat{X}_j(t)|\mathbf{M_{X_i}}$, the cross-mapped estimation of $X_j(t)$, is generated by identifying $E+1$ neighbors in $X_j$ mapped from the $E+1$ nearest neighbors in the $\mathbf{M_{X_i}}$ and computing the weighted mean of these $E+1$ values. $E$ represents the dimension of the reconstructed space. In this example, the original system is $N$-dimensional, so $E$ should be smaller than $2N+1$ ($E \leq 2N+1$) based on Whitney embedding theorem [37].

$$\hat{X}_j(t)|\mathbf{M_{X_i}} = \sum w_k X_j(t_k) \qquad k = 1, 2, \ldots, E+1 \qquad (13)$$

where $w_k$ is the weight concerning the distance between $x_i(t)$ and its $k$-th nearest neighbor on $\mathbf{M_{X_i}}$ and $X_j(t_k)$ are the instantaneous values of $X_j$. $w_k$ is defined by

$$w_k = u_k / \sum u_l \qquad l = 1, 2, \ldots, E+1 \qquad (14)$$

where

$$u_k = exp\{-d[x_i(t), x_i(t_k)]/d[x_i(t), x_i(t_1)]\} \qquad (15)$$

and $d[x, y]$ represents the Euclidean distance between two vectors.

Next, the correlation between $X_j$ and $\hat{X}_j(t)|\mathbf{M_{X_i}}$ is calculated, and the procedures are iterated by increasing the length $L$. If $X_i$ and $X_j$ are dynamically coupled and $X_i$ influences $X_j$, $\hat{X}_j(t)|\mathbf{M_{X_i}}$ should converge to $X_j(t)$, which represents that the last $n$ values of the correlation coefficients should maintain stability.

We use the CCM implementation of the Python library skccm [6] in this study.

CCM performs strongly in coupled systems for detecting causal relationships in coupled time series; however, it may miss simple linear causal relationships and can be sensitive to noise. The other three complementary base algorithms, GC, NTE, and CCM in our ensemble model, are good supplements to improve the robustness and reliability of causality detection. These algorithms are well-suited for detecting linear causal relationships, processing non-linear causality, and handling large-scale complex systems, respectively.

## 3. A Two-Phase Multi-Split Causal Ensemble Model

We propose to combine the methods introduced in the previous section in a two-phase ensemble learning algorithm for time series causal reasoning. The general procedure of the ensemble method is introduced in Section 3.1, the details of the key components are discussed in Sections 3.2 to 3.4, and the evaluation processing is covered in Section 3.5.

### 3.1. General Pipeline

To improve the accuracy of existing causal inference approaches, we propose a two-phase multi-split causal ensemble model to derive the full benefit of the four causal inference algorithms (GC, NTE, PCMCI+, and CCM). In many experiments [38,39], multiple algorithms often lead to divergent causality conclusions when processing the same dataset, which affects the trustworthiness of causal inference results. The proposed ensemble framework aims to produce more robust results by integrating multiple causal detectors. Sources of noise can often temporarily affect time series, leading to data which are unrepresentative of the underlying causal mechanisms in sections of the time series. This can lead to wrongly inferred causal relationships if the time series is examined as a whole.

To circumvent the mentioned challenges, our model follows a split-and-ensemble approach as Figure 1 shows, meaning we first partition the given data, apply causal inference methods to each partition, and, finally, combine the results. The split-and-ensemble process makes full use of the data and reduces the impacts of noise in the datasets. In the data partitioning step, the time series are split into many partitions with an overlapping partitioning process. After each base algorithm has been applied to each partition, the first ensemble phase uses a GMM to integrate the causal results of each base algorithm from different data partitions in order to improve the robustness of each base algorithm. This leads to one combined causal inference result for each base algorithm, and the trustworthiness assessment is conducted to check the stability of the result. Therefore, the first-layer ensemble outputs four combined causal inference results associated with the base learners, as well as their evaluation results, which is demonstrated in the GMM ensemble phase of Figure 1. The second ensemble phase combines these results through three decision-making rules considering the first-layer evaluation results, which take advantage of the strengths of diverse causal inference algorithms and make the results more robust.

Finally, the ensemble result is optimized by removing the indirect causal links and the credibility of the final result is evaluated by a proposed assessment coefficient. The overall process is illustrated in Figure 1 and formalized in Algorithm 1. The following sections will explain the individual steps of the algorithm which have been briefly introduced above.

### 3.2. Data Partitioning

Before the proposed ensemble algorithm can be applied, the time series observations are split into several partitions to be processed by the four base learners. If there is any seasonality or cyclicity in the time series, observations should be divided to account for these properties. For example, if a process has a cycle of two hours, the time series describing this process can be divided into the groups zero to four hours, two to six hours, four to eight hours, and so on. Otherwise, the length of each slice is flexible. In the partitioning process, one subset is overlapped with part of another one. It means some data instances are sampled more than once, which allows the information of the data to be reused in different partitions and enhances the stability of the data-driven model. A brief example of data splitting is demonstrated in Figure 2.

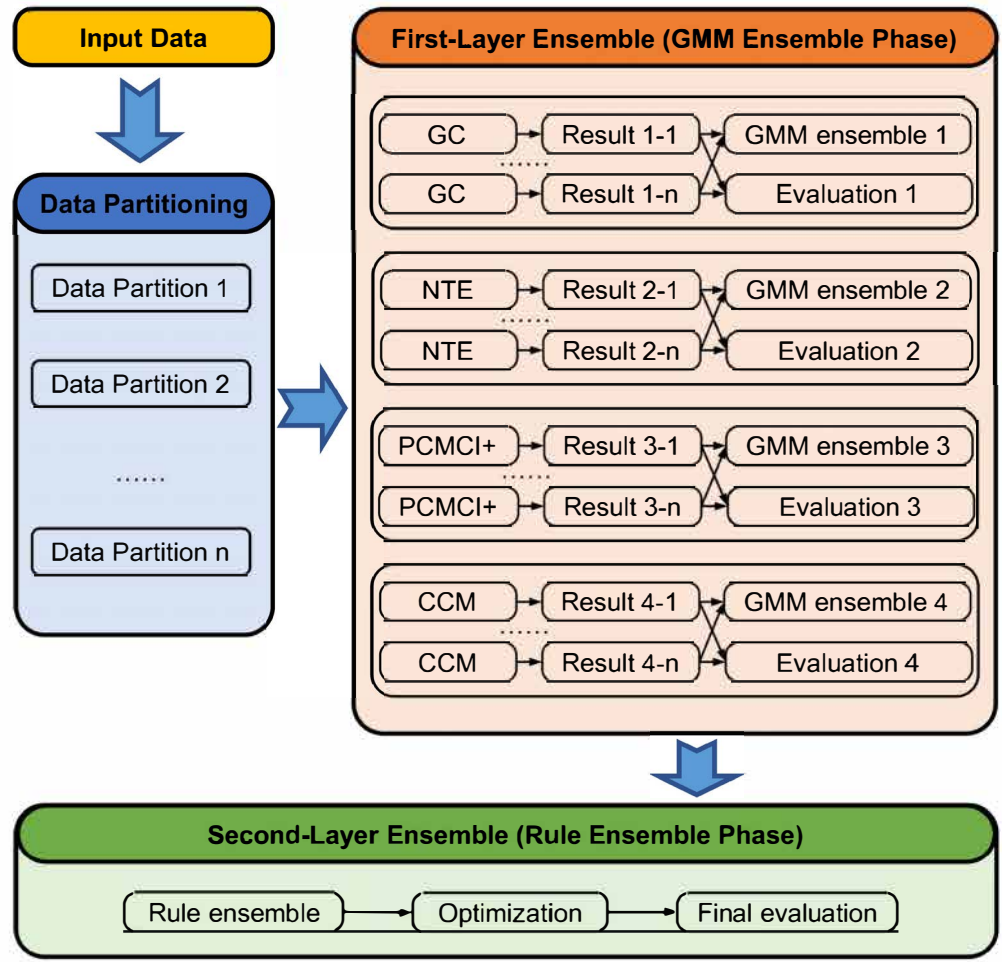

Figure 1. A two-phase multi-split causal ensemble framework.

Algorithm 1 demonstrates the general pipeline of the proposed causal ensemble model, including Algorithms which are explained in detail in the following sections.

---

**Algorithm 1** The data-driven two-phase multi-split causal ensemble model

---

**Input:** Time series dataset $\mathbf{X} = (X^1, X^2, \ldots, X^N)$, maximum time lag $\tau_{\max}$, the number of data partitions $M$.
**Output:** Causal strength matrix $\widetilde{\mathbf{MRE}}$ with the size $N \times N$, credibility score $CS$, causal graph $\mathcal{G}$
1: **for all** $m \in \{1, \ldots, M\}$ **do**
2:    Run GC, NTE, PCMCI+, and CCM in parallel    ▷ Base learners
3:    Return strength matrices $\mathbf{M}_{l,m}$ $(l = 1, 2, 3, 4)$
4: **end for**
5: $\mathbf{ME}_l, \mathbf{T}_l \leftarrow$ Algorithm 2 $(\mathbf{M}_{l,1}, \ldots, \mathbf{M}_{l,M})$ $(l = 1, \ldots, 4)$    ▷ GMM ensemble phase (Section 3.3)
6: $\mathbf{MRE} \leftarrow$ Algorithm 3 $(\mathbf{ME}_1, \ldots, \mathbf{ME}_4; \mathbf{T}_1, \ldots, \mathbf{T}_4)$    ▷ Rule ensemble phase (Section 3.4)
7: $\widetilde{\mathbf{MRE}} \leftarrow$ optimization($\mathbf{MRE}$)    ▷ Model optimization (Section 3.5)
8: Obtain the causal graph $\mathcal{G}$ concerning $\widetilde{\mathbf{MRE}}$
9: Compute $CS$ by Equation (26)

---

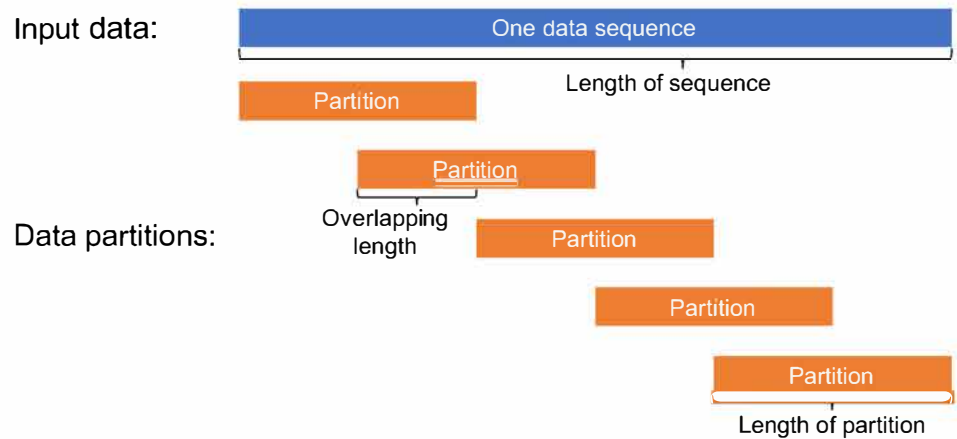

**Figure 2.** Illustration of data partitioning into overlapping subsets.

*3.3. GMM Ensemble Phase*

After the partitioning, the four base learners are applied to process each data partition separately and in parallel. After partitioning of the data into $n$ partitions, each partition is processed by the four individual models, and $4 \times n$ causal strength matrices are obtained (see Figure 1). Next, the results from each individual causal inference method are combined using a Gaussian mixture model (GMM).

When using GMMs for clustering, each Gaussian component represents a cluster, and the corresponding weight represents the probability that a data point belongs to said cluster [40]. A datum point is assigned to the cluster with the highest probability, which is called the soft assignment. Unlike the k-means algorithm, which can only detect spherical clusters, the GMM can process elliptical or oblong clusters. Moreover, in contrast to the soft assignments of GMM clustering, the k-means algorithm can only perform hard assignments. Since the transformed dataset at this stage is oblong in shape, GMM clustering is the more suitable approach to ensemble causal models. In this ensemble learning framework, the relationships between two variables are divided into two groups, causal and non-causal. Therefore, the number of components of the GMM is set as two.

In this ensemble phase, a two-component GMM is fitted to ensemble the causal strength matrices from the same causal detector but from different data partitions. Figure 3 visualizes the steps of processing the observations in this phase in a dataset with $n$ features being split into $K$ partitions. In this example, the results are $K$ causal strength matrices with sizes that are all $n \times n$ after the processing of any of the four base learners. The $\mathbf{M}_k(i,j)$ ($k \in \{1,\ldots,K\}$)($i,j \in \{1,\ldots,n\}$) represents the causal strength from feature $i$ to feature $j$ for the $k$-th partition, where $\mathbf{M}_k(i,j) \in [0,1] \forall i,j,k$. Before the post-processing, the matrices are filtered in the first step as Equation (16) shows, because the causal strengths in the four base learners are all based on the correlation coefficient (CC). In order to not be too restrictive too early, two features are considered a low-correlation pair and the causal inference is weakly trustworthy if $CC \leq 0.3$. Equation (16) shows the processing of weakly trustworthy causal links.

$$\mathbf{M}_k(i,j) \leftarrow 0 \quad if \ \mathbf{M}_k(i,j) \leq 0.3 \tag{16}$$

In the second step, all the $K$ matrices are flattened to vectors with size $n(n-1)$, where $n$ self-adjacent values are removed. In other words, the diagonal of the matrix is removed because this model solely analyzes the relationships between different variables. In the time series, the future values of a feature are assumed to be dependent on the past values, so there is a causal relationship within the feature. However, it is not shown in the causality model as the model is the rolled-up version without unrolling in the time dimension. Then, the data points with the same position indices are integrated into one data instance with $K$ dimensions. Thus, a new dataset $M^*$ containing $n(n-1)$ $K$-dimensional instances is created as the input of the GMM clustering, which is illustrated in Figure 3 (step 2). The left numbers are the indices of the instance in the $\mathbf{M}^*$, and the right tuples represent the position of the $\mathbf{M}_k$ where the values are generated. For instance, the first instance in the $\mathbf{M}^*$ is $[\mathbf{M}_1(1,2), \mathbf{M}_2(1,2), \ldots, \mathbf{M}_K(1,2)]$. The $m$-th instance in the $\mathbf{M}^*$ is labeled $\mathbf{M}^*[m](i,j)$, where $m$ is the left index and $(i,j)$ refers to the right index in Figure 3 (step 2).

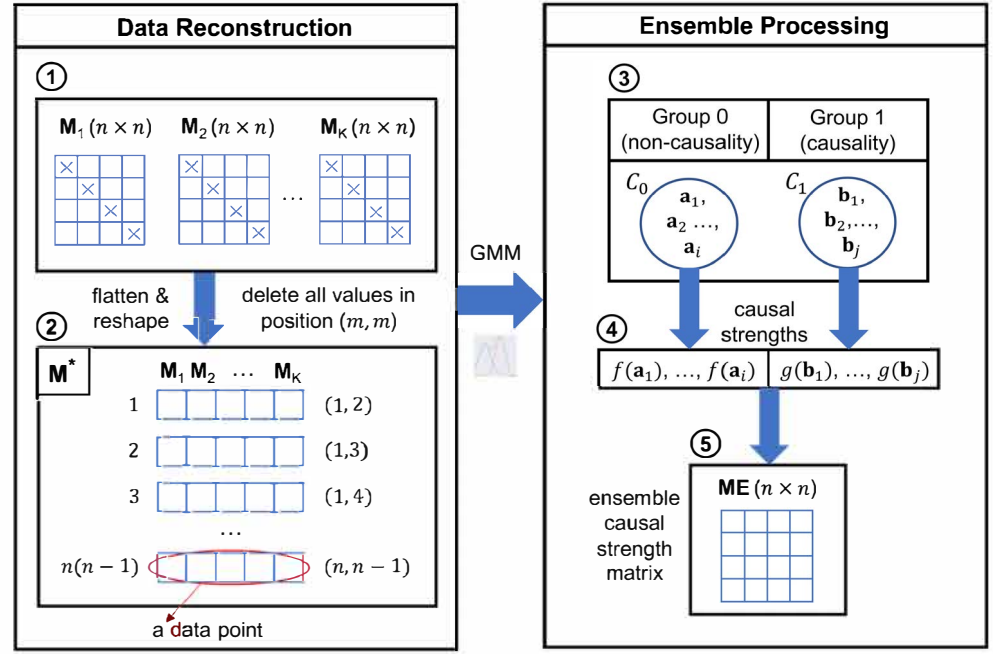

**Figure 3.** Illustration of GMM ensemble phase.

Next, in the third step, the GMM with two components is fitted to cluster the input data. The Gaussian with the lower mean is denoted as the cluster for the causal strength data from the feature pairs without causal relationships, which is labeled "Group 0". The Gaussian with the higher mean represents the cluster containing the strength values of causal pairs, which is labeled "Group 1". All of the $n(n-1)$ inputs are assigned to the two groups by the fitted GMM. Finally, each clustered $K$-dimensional data instance is mapped to a single floating point value. In Figure 3 (step 3), $C_0 = \{\mathbf{a}_1, \ldots, \mathbf{a}_i\}$ is the set containing the instances in "Group 0" and $f(x)$ is the corresponding mapping function.

$$f(\mathbf{a}_l) = 0 \qquad \forall \mathbf{a}_l \in C_0 \tag{17}$$

$C_1 = \{\mathbf{b}_1, \ldots, \mathbf{b}_j\}$ is the set including the data points in "Group 1" and the corresponding mapping function is $g(x)$.

$$g(\mathbf{b}_l) = median^+(\mathbf{b}_l) \qquad \forall \mathbf{b}_l \in C_1 \tag{18}$$

where $median^+(\mathbf{x})$ is a function to calculate the median of non-zero values in a vector. For instance, assume $\mathbf{b}_l = [x_1, x_2, \ldots, x_K]$ $\forall \mathbf{b}_l \in C_1$, and $\mathcal{X}_l = \{x_1, x_2, \ldots, x_K\}$ is the set containing the values of the coordinate of the vector $\mathbf{b}_l$. Define $\mathcal{X}_{l2} \subseteq \mathcal{X}_l$ by removing all elements valued 0.

$$median^+(\mathbf{b}_l) = median(\mathcal{X}_{l2}) \qquad \forall \mathbf{b}_l \in C_1 \tag{19}$$

where $median(\mathcal{X}_{l2})$ denotes the median of the values in the set $\mathcal{X}_{l2}$. In this instance, the median is selected instead of the mean because it represents the values' average information without the influence of outliers.

After the processing, an ensemble causal strength matrix $\mathbf{ME}$ is produced separately for each base learner, and its size is $n \times n$, which is shown in Figure 3 (step 5). The values of $\mathbf{ME}(i,i)$ $\forall i \in \{1,\ldots,n\}$ are set at 0 because causal detection does not consider self-related pairs as discussed before. The values of $\mathbf{ME}(i,j)$ $\forall i,j \in \{1,\ldots,n\}$ ($i \neq j$) rely on the ensemble processing and correspond to the instance $\mathbf{M}^*[m](i,j)$ (Figure 3 (step 2)), which is processed following Equation (20).

$$\mathbf{ME}(i,j) = \begin{cases} f(\mathbf{M}^*[m](i,j)) & \mathbf{M}^*[m](i,j) \in C_0 \\ g(\mathbf{M}^*[m](i,j)) & \mathbf{M}^*[m](i,j) \in C_1 \end{cases} \quad (20)$$

where $f(x)$ is based on Equation (17) and $g(x)$ is derived from Equation (18).

It is necessary to evaluate the trustworthiness of the causal relationships from the GMM ensemble phase and make preparations for the next ensemble phase in Section 3.3. To accomplish that goal, an assessment pipeline is designed, as Figure 4 illustrates.

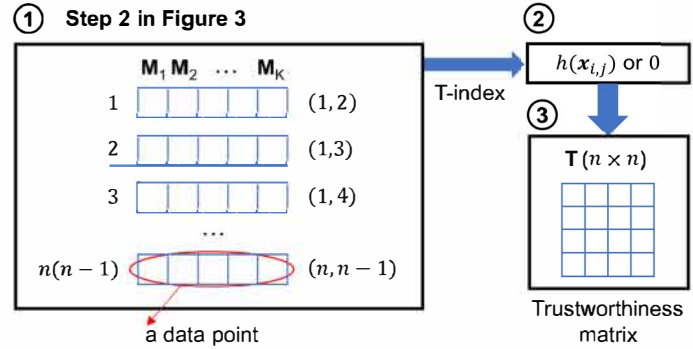

**Figure 4.** Trustworthiness evaluation for GMM ensemble.

The trustworthiness matrix $\mathbf{T}$ with the size $n \times n$ is proposed corresponding to the ensemble causal strength matrix $\mathbf{ME}$. The elements are then computed by the trustworthiness index in Equations (21) and (22). If the value of $\mathbf{ME}(i,j)$ is 0, the corresponding element $\mathbf{T}(i,j)$ is set at 0 as well. This is because, under the requirements of the following rule ensemble phase, only causal relationships require assessment. If the value of $\mathbf{ME}(i,j)$ is not 0, the value of $\mathbf{T}(i,j)$ is set by Equation (22).

$$\mathbf{T}(i,j) = \begin{cases} 0 & \mathbf{ME}(i,j) = 0 \\ h(\mathbf{x}_{i,j}) & \mathbf{ME}(i,j) \neq 0 \end{cases} \quad (21)$$

where

$$h(\mathbf{x}_{i,j}) = \frac{mean^*(\mathbf{x}_{i,j}) \times m}{[std^*(\mathbf{x}_{i,j}) + \delta] \times K} \quad (22)$$

where $\mathbf{x}_{i,j}$ is a variable containing $K$ elements. This variable is transformed from the $K$-dimensional data point in the dataset of step 2 in Figure 4, which is derived from the elements $\mathbf{M}_k(i,j)$, where $k = 1,\ldots,K$. $mean^*(\mathbf{x}_{i,j})$ denotes the mean of non-zero values in $\mathbf{x}_{i,j}$, $std^*(\mathbf{x}_{i,j})$ represents the standard deviation of non-zero values in $\mathbf{x}_{i,j}$, $m$ refers to the number of the non-zero values, and $K$ is the number of elements in $\mathbf{x}_{i,j}$. $\delta$ is an extremely small value (e.g., $10^{-20}$) used to avoid a divisor being equal to zero.

The range of any $\mathbf{T}(i,j)$ is $[0,+\infty)$. A value approaching 0 points to barely detectable causal relationships, whereas an extremely high value represents stable and strong causal relationships across all data partitions. In Equation (22), a high mean represents a strong causal relationship. A high parameter $m$ indicates a good consistency in the causal detection among different partitions. A low standard deviation indicates the stability of causal inference because the values are distributed over a narrow range. Therefore, a higher value of $h(x)$ indicates greater trustworthiness in the GMM-ensemble processing.

Algorithm 2 outlines the procedure of the GMM ensemble phase.

**Algorithm 2** GMM ensemble phase

**Input:** $4 \times K$ causal strength matrices $\mathbf{M}_{l,k}$ with the size $n \times n$ (4 refers to the four base learners, $K$ is the number of partitions, $l = 1,\ldots,4$ and $k = 1,\ldots,K$)
**Output:** GMM-ensemble strength matrices $\mathbf{ME}_l$ with the size $n \times n$ and trustworthiness matrices $\mathbf{T}_l$ with the size $n \times n$ ($l = 1,2,3,4$)

```
1:  for all l in {1,2,3,4} do                                         ▷ Run in parallel
2:      for all k in {1,...,K} do
3:          flatten M_{l,k} and remove values M_{l,k}(i,i) (i = 1,...,n) to the size 1 × n(n − 1)
4:      end for
5:      F_l ← combine all flattened M_{l,k}            ▷ the size of F_l is K × n(n − 1)
6:      (Group 0, Group 1) ← Run a GMM (number of components = 2) to process F_l
7:      for all x_{i,j} in F_l do
8:          if x_{i,j} is in Group 0 then
9:              ME_l(i,j) ← 0
10:             T_l(i,j) ← 0
11:         else
12:             ME_l(i,j) ← g(x_{i,j})                                 ▷ Equation (18)
13:             T_l(i,j) ← h(x_{i,j})                                  ▷ Equation (22)
14:         end if
15:     end for
16: end for
```

*3.4. Rule Ensemble Phase*

The GMM ensemble phase is conducted to process the four base learners separately and in parallel. The outputs of this step are the four ensemble causal strength matrices $\mathbf{ME}_1$, $\mathbf{ME}_2$, $\mathbf{ME}_3$, and $\mathbf{ME}_4$, and the corresponding trustworthiness matrices $\mathbf{T}_1$, $\mathbf{T}_2$, $\mathbf{T}_3$, and $\mathbf{T}_4$ with sizes that are all $n \times n$. These serve as the inputs of the rule ensemble phase, which is described in the following.

In this phase, the four intermediate ensemble matrices are integrated into one rule-ensemble matrix $\mathbf{MRE}$ with the size $n \times n$ based on three rules as follows. To determine the element $\mathbf{MRE}(i,j)$ ($i,j = 1,2,\ldots,n$), the set $\mathcal{R}_{i,j} = \{\mathbf{ME}_1(i,j), \mathbf{ME}_2(i,j), \mathbf{ME}_3(i,j), \mathbf{ME}_4(i,j)\}$ is created, and $r_{i,j}$ is denoted as the number of non-zero elements in $\mathcal{R}_{i,j}$. Before conducting the following rules, the $\mathbf{ME}$ is filtered to improve the credibility of the ensemble processing. Through a number of experiments, for any $(i,j)$, if $\mathbf{T}_k(i,j) < 1.0$ ($k = 1,2,3,4$), the corresponding element in the causal strength matrix $\mathbf{ME}_k(i,j)$ is set at 0. This highlights that such detected cause–effect pairs can not be trusted.

Next, three ensemble rules are developed to combine the intermediate ensemble results. The formal definition is as follows. Rule 1 states that there is no causal link if more than half of the intermediate ensemble results indicate non causality. Similarly, as Rule 3 shows, when the majority of the intermediate results specify the causal relationship, the causality is confirmed. Rule 2 expresses the decision-making process without majority. The decisions are made based on the trustworthiness matrices and corresponding thresholds. Furthermore, the quantitative causality result, causal strength, is determined by a weighting strategy.

**Rule 1:** When $r_{i,j} = 0$ or $1$, set $\mathbf{MRE}(i,j)$ at $0$.

**Rule 2:** When $r_{i,j} = 2$, suppose $\mathbf{ME}_{k1}(i,j) > 0$ and $\mathbf{ME}_{k2}(i,j) > 0$.
If $\max\{\mathbf{T}_{k1}(i,j), \mathbf{T}_{k2}(i,j)\} > \alpha_{21}$, set $\mathbf{MRE}(i,j)$ at $WCS(i,j)$.
If $\max\{\mathbf{T}_{k1}(i,j), \mathbf{T}_{k2}(i,j)\} < \alpha_{21}$ and $\min\{\mathbf{T}_{k1}(i,j), \mathbf{T}_{k2}(i,j)\} > \alpha_{22}$, set $\mathbf{MRE}(i,j)$ at $WCS(i,j)$ as well, otherwise set $\mathbf{MRE}(i,j)$ at $0$.

**Rule 3:** When $r_{i,j} = 3$ or $4$, set $\mathbf{MRE}(i,j)$ at $WCS(i,j)$.

The parameters $\alpha_{21}$ and $\alpha_{22}$ are the relevant threshold. Based on many experiments, $\alpha_{21}$ can be selected between 10 and 20, and $\alpha_{22}$ can be chosen between 1.5 and 2.5.

$WCS(i, j)$ is the weighted causal strength based on Equation (23).

$$WCS(i,j) = \sum_{k=1}^{4} \widetilde{\mathbf{T}}_k(i,j)\mathbf{ME}_k(i,j) \qquad \forall i,j = 1,\ldots,n \qquad (23)$$

where $\widetilde{\mathbf{T}}_k(i,j)$ refers to the normalized weight based on the trustworthiness matrix $\mathbf{T}_k$ ($k = 1, 2, 3, 4$), which ranges from 0 to 1. Furthermore, the $\mathbf{ME}_k$ ($k = 1, 2, 3, 4$) represents the values in the GMM-ensemble causal strength matrix.

$$\widetilde{\mathbf{T}}_k(i,j) = \frac{\mathbf{T}_k(i,j)}{\sum_{l=1}^{4} \mathbf{T}_l(i,j)} \qquad \forall i,j = 1,\ldots,n \qquad (24)$$

Algorithm 3 demonstrates the procedure of the rule ensemble phase.

---

**Algorithm 3** Rule ensemble phase

**Input:** GMM-ensemble strength matrices $\mathbf{ME}_l$ with the size $n \times n$ and trustworthiness matrices $\mathbf{T}_l$ with the size $n \times n$ ($l = 1,\ldots,4$), the threshold $\alpha_{21}$ and $\alpha_{22}$
**Output:** GMM-ensemble strength matrices $\mathbf{MRE}$ with the size $n \times n$
1: Initialize $\mathcal{R}_{i,j} = \{\mathbf{ME}_1(i,j), \mathbf{ME}_2(i,j), \mathbf{ME}_3(i,j), \mathbf{ME}_4(i,j)\}$ $\forall i,j \in \{1,\ldots,n\}$ & $i \neq j$ and $r_{i,j}$ representing the number of non-zero elements in $\mathcal{R}_{i,j}$
2: **for all** $\mathbf{ME}_l(i,j)$ $\forall l \in \{1,\ldots,4\}$ $i,j \in \{1,\ldots,n\}$ & $i \neq j$ **do**
3:     **if** $\mathbf{T}_l(i,j) < 1$ **then**
4:         $\mathbf{ME}_l(i,j) \leftarrow 0$
5:     **end if**
6: **end for**
7: **for all** $\mathcal{R}_{i,j}$ $\forall i,j \in \{1,\ldots,n\}$ & $i \neq j$ **do**
8:     **if** $r_{i,j} = 0$ or $1$ **then**     ▷ **Rule 1**
9:         $\mathbf{MRE}(i,j) \leftarrow 0$
10:     **else if** $r_{i,j} = 2$ **then**     ▷ **Rule 2**
11:         Select the largest element in $\mathcal{R}_{i,j}$ $\mathbf{ME}_{k1}(i,j)$ and the second largest one $\mathbf{ME}_{k2}(i,j)$
12:         **if** $\mathbf{T}_{k1}(i,j) \geq \alpha_{21}$ **then**
13:             $\mathbf{MRE}(i,j) \leftarrow WCS(i,j)$     ▷ Computing $WCS$ in Equation (23)
14:         **else if** $\mathbf{T}_{k1}(i,j) < \alpha_{21}$ & $\mathbf{T}_{k2}(i,j) \geq \alpha_{22}$ **then**
15:             $\mathbf{MRE}(i,j) \leftarrow WCS(i,j)$
16:         **else**
17:             $\mathbf{MRE}(i,j) \leftarrow 0$
18:         **end if**
19:     **else**     ▷ **Rule 3**
20:         $\mathbf{MRE}(i,j) \leftarrow WCS(i,j)$
21:     **end if**
22: **end for**

---

The three rules constitute the rule ensemble phase, where the causal pairs detected by multiple base learners with high trustworthiness scores are selected and embedded into the final causal strength matrix. The comprehensive evaluation and selection process enhances the reliability of the causal models and is more robust than majority voting and averaging.

### 3.5. Model Optimization

In the rule-ensemble result $\mathbf{MRE}$, both direct and indirect causal relationships are present. In the applications, the direct causal relationships are more significant than the indirect ones for analyzing the system's performance [41]. Hence, the indirect causal links should be extracted and removed from the $\mathbf{MRE}$.

Many experiments [41] demonstrate that the indirect adjacency is weaker than the direct one. As Figure 5 shows, for instance, $X^2$ influences $X^3$ directly, but $X^1$ can only influence $X^3$ through $X^2$. $\mathbf{MRE}(i,j)$ denotes the causal strength from $X^i$ to $X^j$. Under the conclusion from the experiments, for instance, the triple $\langle X^1, X^2, X^3 \rangle$ contains the inequalities $\mathbf{MRE}(1,2) > \mathbf{MRE}(1,3) > 0$ and $\mathbf{MRE}(2,3) > \mathbf{MRE}(1,3) > 0$. $\mathbf{MRE}(1,4)$ is the weakest value among all causal strengths.

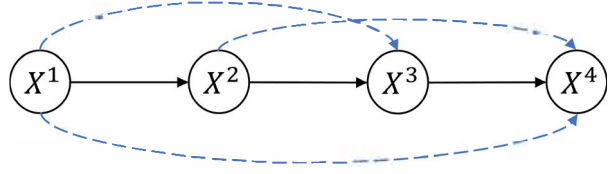

**Figure 5.** Direct and indirect causal links. $X^i (i = 1, \ldots, 4)$ in black circles represent four distinct time series. The blue dashed lines denote the indirect links, the black solid lines represent the direct adjacencies, and the arrows illustrate the directions of the cause–effect relationships.

To remove the indirect links of the time series dataset $\mathbf{X} = (X^1, X^2, \ldots, X^N)$, all triples $\langle X^i, X^j, X^k \rangle$ $(i, j, k = 1, \ldots, N, i \neq j \neq k)$ meeting the conditions of $(X^i \to X^j \to X^k)$ and $(X^i \to X^k)$ in the rule-ensemble result are analyzed. If $\mathbf{MRE}(i,j) > \mathbf{MRE}(i,k) > 0$ and $\mathbf{MRE}(j,k) > \mathbf{MRE}(i,k) > 0$, it indicates that $(X^i \to X^k)$ is the indirect adjacency and should be removed from the causal links. In turn, this means that $\mathbf{MRE}(i,k)$ should be set at 0. The filtered matrix from $\mathbf{MRE}$ is denoted as $\widetilde{\mathbf{MRE}}$.

The optimization step marks the end of the causal discovery procedure. The final causal strength matrix $\widetilde{\mathbf{MRE}}$, as well as the corresponding causal graph $\mathcal{G}$ is then produced. In the next section, an evaluation index is proposed to assess the credibility of the ensemble model.

## 4. Causal Ensemble Model Evaluation

When evaluating the causal results, the most commonly used approach in applications is the confusion matrix, which compares the detected causal links with the ground truth [42]. However, no ground truth is provided in many cases, which makes it impossible to evaluate causal results by comparing them with true causal relationships. Therefore, it is necessary to explore an evaluation metric that is based on the mechanisms of the algorithms. Such a metric should be able to statistically evaluate the robustness and stability of the algorithm in various experiments. When evaluating the proposed causal ensemble model, particular attention should be given to assessing the ensemble process.

In this section, the credibility score (CS) is developed based on the final causal strength matrix $\widetilde{\mathbf{MRE}}$ and the first-layer GMM-ensemble causal strength matrices $\mathbf{ME}_k$ ($k = 1, 2, 3, 4$) to evaluate the credibility of the proposed causal ensemble model. This evaluation metric is to measure the agreement between the intermediate ensemble parts. When most of the base learners reach similar results, it represents that the GMM ensemble phase successfully reduces the influence of noise and we define the similarity of the four intermediate causal results after the GMM processing as stable. Equation (25) defines the similarity index for a single causal relationship between two variables and Equation (26) denotes the credibility score $CS$ by averaging the similarity indexes and normalizing it to the range $[0, 1]$. $CS$ represents the stability and credibility of the causal results, because the more stable the results from the base learners, the more trustworthy the causal results in the rule ensemble phase.

Assume that the input contains $n$ features, for all $i, j = 1, \ldots, n$ and $i \neq j$,

$$CS_{i,j} = \begin{cases} U/K & \widetilde{\mathbf{MRE}}(i,j) = 0 \\ (K-U)/K & \widetilde{\mathbf{MRE}}(i,j) \neq 0 \end{cases} \qquad (25)$$

where $K$ is the number of base learners ($K = 4$ in this ensemble model), and $U$ represents the number of zero elements in the set $\{\mathbf{ME}_1(i,j), \mathbf{ME}_2(i,j), \mathbf{ME}_3(i,j), \mathbf{ME}_4(i,j)\}$.

The similarity index $CS_{i,j}$ assesses the consistency between the final result of the causal ensemble model and the four results obtained from the base algorithms applying the GMM ensemble. $K$ and $U$ are utilized to quantify the consistency in Equation (25). Based on rule 1–3 in rule ensemble phase (Section 3.4), if $\widetilde{MRE}(i,j) = 0$, $U \in \{2,3,4\}$. If $\widetilde{MRE}(i,j) \neq 0$, $U \in \{0,1,2\}$. Hence, all the potential values of $CS_{i,j}$ are included in $\{0.5, 0.75, 1.0\}$.

Then, the mean of all $CS_{i,j}$ values are calculated, and that mean number is normalized to the range of $[0,1]$, which is the final evaluation index $CS$. Normalization makes the metric comparable across indicators, because all indexes in the confusion matrix evaluating the causal results are in the range $[0,1]$.

$$CS = \left( \frac{\sum CS_{i,j}}{n(n-1)} - 0.5 \right) \times 2 \qquad (26)$$

If $CS < 0.5$, it indicates that most $CS_{i,j}$ values are 0.5 in the assessment. This reveals that, in the detection of these causal pairs, half of the base learners affirm the causal relationships while the other half do not. This is the maximum level of disagreement that is possible among parts of the ensemble. When such a credibility score is observed, the results should, therefore, be treated with caution. In contrast, when the values 0.75 and 1.0 are more prevalent, there is more agreement among parts of the ensemble. With $CS = 1.0$, all of the base algorithms generate the same causal results, increasing the confidence in the inferred causal relationships. Between the extremes of $CS < 0.5$ and $CS = 1.0$, there is a continuum of values quantifying the level of agreement between different parts of the ensemble, which can be used to inform ones confidence in the results. Table 1 shows the relationships between the levels of credibility and the range of $CS$.

Table 1. Level of credibility of the causal ensemble model.

| Range of $CS$ | Level of Credibility |
| --- | --- |
| $0 \leq CS < 0.5$ | weak credibility |
| $0.5 \leq CS < 0.75$ | medium credibility |
| $0.75 \leq CS < 1$ | strong credibility |

This proposed $CS$ index is designed specifically for the causal ensemble model in this study. This is because all of the components used to calculate the evaluation metric are based on specific parameters within this model.

## 5. Experiments and Discussion
### 5.1. Set Up

A series of experiments are designed and conducted in our study to compare the performance of the ensemble causal results and the results of the individual base algorithms, testing whether the proposed causal ensemble model outperform its base learners. We use generated datasets with distinct sizes and complexities in the experiments to test how the data volumes and non-linear relationships affect the performance of the proposed model.

The following parameters are set when initializing the algorithm to process distinct datasets, and they are selected based on several experiments. The global parameters are the number of data partitions $K$ and the maximum time lag $\tau_{max}$, which are set depending on the input data. In the GC and PCMCI+, the $p$-value thresholds for significant tests are both set at 0.05. The following parameters in the NTE and CCM are selected to keep a balance of runtime and accuracy. In the NTE, the number of nearest neighbors is set to 6. In the CCM, the percentage of training data is set at 0.75, and the number of iterations is 25. Moreover, the number of values for checking convergence is set to 6 and the threshold of determining convergence is set to 0.03. In the GMM ensemble phase, the number of initializations for the GMM is set to 10 to select a reasonable clustering result and the other parameters of the GMM are considered the default values in the Python library Scikit-learn (version 1.1.2) [43]. In the rule ensemble phase, the two thresholds in "Rule 2" are set to 10.0 and 2.0, respectively, based on the discussion in Section 3.3.

*5.2. Description of the Datasets*

For the experiments examining the synthetic data, three datasets with different sizes and complexities are generated based on predefined equations, similarly to [44]. In the generated datasets, $X_t^i$ denotes the $i$-th variable at time $t$, and $\epsilon$ is the Gaussian-distributed random noise following $\mathcal{N}(0,1)$ with distinct random seeds. datasets 1–3 are demonstrated in Equations (27)–(29), respectively.

Dataset 1 is a linear dataset containing 5 variables. In total, 20,000 observations are generated from each variable and the maximum time lag $\tau_{\max}$ is 4. Variables $X^1$ and $X^4$ are generated following the uniform distribution $\mathbf{U}[0,100]$ with different random seeds, which indicates that $X^1$ and $X^4$ are independent of each other. $X^2$, $X^3$, and $X^5$ are computed based on Equation (27). The coefficients are selected randomly around $[0.5, 1.5]$ to control the range of each sequence to avoid the exponential trend. The random selection of coefficients can test the stability of the proposed framework, which can be applied to a diverse range of time series datasets flexibly.

$$\begin{cases} X_t^2 = 1.34 X_{t-2}^1 - 1.23 X_{t-4}^2 + \epsilon_{11} \\ X_t^3 = -1.26 X_{t-1}^2 + 1.03 X_{t-2}^2 + \epsilon_{12} \\ X_t^5 = 0.71 X_{t-2}^4 + 1.05 X_{t-3}^4 + \epsilon_{13} \end{cases} \quad (27)$$

Dataset 2 is a non-linear dataset involving 5 variables. In total, 30,000 observations are generated from each variable and some of the variables are interacting. The maximum time lag $\tau_{\max}$ is 5. The variables $X^1$ and $X^4$ are generated following the uniform distribution $\mathbf{U}[0,100]$ with different random seeds, which indicates that $X^1$ and $X^4$ are independent of each other. $X^2$, $X^3$, and $X^5$ are computed based on Equation (28). The coefficient selection follows the principle in dataset 1.

$$\begin{cases} X_t^2 = 1.06 X_{t-1}^2 - 1.22 X_{t-3}^1 X_{t-2}^2 - 1.41 (X_{t-3}^1)^2 X_{t-4}^1 + \epsilon_{21} \\ X_t^3 = -1.23 X_{t-1}^2 X_{t-2}^2 X_{t-2}^3 + 0.69 (X_{t-1}^2)^2 + 1.07 (X_{t-2}^2)^2 X_{t-5}^2 X_{t-3}^3 + \epsilon_{22} \\ X_t^5 = -0.78 X_{t-4}^1 X_{t-2}^4 + 0.91 (X_{t-1}^4)^2 X_{t-4}^5 - 0.86 (X_{t-1}^1)^2 X_{t-2}^5 + 1.17 X_{t-5}^1 X_{t-3}^4 + \epsilon_{23} \end{cases} \quad (28)$$

Dataset 3 is a more complex non-linear dataset including 12 variables, where 45,000 observations are generated from each variable and $\tau_{\max}$ is 6. The variables $X^1$, $X^4$, and $X^9$ are generated following the uniform distribution $\mathbf{U}[0,100]$ with different random seeds, which indicates that $X^1$, $X^4$, and $X^9$ are independent of each other. $X^2$, $X^3$, $X^5$, $X^6$, $X^7$, $X^8$, $X^{10}$, $X^{11}$, and $X^{12}$ are computed based on Equation (29). The coefficient selection follows the principle in dataset 1.

$$\begin{cases} X_t^2 = 0.99 X_{t-3}^1 X_{t-2}^2 - 1.24 X_{t-2}^1 (X_{t-4}^1)^2 + \epsilon_{31} \\ X_t^3 = -1.06 X_{t-2}^9 X_{t-2}^3 + 0.54 (X_{t-1}^9)^2 + 0.72 (X_{t-5}^9)^2 X_{t-3}^3 + \epsilon_{32} \\ X_t^5 = -0.86 X_{t-6}^1 + 0.67 X_{t-1}^4 - 0.88 X_{t-2}^1 X_{t-2}^4 + \epsilon_{33} \\ X_t^6 = 0.69 X_{t-1}^6 (X_{t-2}^8)^2 - 0.59 \sin(X_{t-1}^8) + \epsilon_{34} \\ X_t^7 = 0.91 X_{t-2}^7 (X_{t-3}^9)^2 + 0.66 X_{t-4}^7 X_{t-4}^9 - 0.33 \exp(X_{t-2}^9) + \epsilon_{35} \\ X_t^8 = 1.18 X_{t-1}^3 - 0.71 \cos(X_{t-3}^3) X_{t-3}^8 + \epsilon_{36} \\ X_t^{10} = 0.78 X_{t-2}^5 X_{t-3}^{10} + 1.02 X_{t-6}^5 + \epsilon_{37} \\ X_t^{11} = 1.31 (X_{t-2}^1)^2 X_{t-4}^{12} + 1.14 (X_{t-2}^{12})^2 X_{t-1}^1 + \epsilon_{38} \\ X_t^{12} = 0.68 X_{t-5}^{10} X_{t-4}^{12} + 0.26 X_{t-2}^9 (X_{t-2}^{10})^2 X_{t-2}^{12} - 0.45 X_{t-3}^9 X_{t-4}^{10} + \epsilon_{39} \end{cases} \quad (29)$$

Table 2 summarizes the general information of the three synthetic datasets and the corresponding data partition settings for the following experiments.

**Table 2.** General information and initialization of the datasets

| Name | Complexity | Variables | Observations | Partitions | Partition Lengths |
|---|---|---|---|---|---|
| dataset 1 | linear | 5 | 20,000 | 10 | 3000 |
| dataset 2 | nonlinear | 5 | 30,000 | 12 | 3750 |
| dataset 3 | extreme nonlinear | 12 | 45,000 | 15 | 4500 |

*5.3. Results*

Figures 6–8 demonstrate the results of processing dataset 1.

In Figure 6, the heatmaps are used to illustrate the causal strength. A dark-blue block in position $(X^i, X^j)$ represents that $X^i$ influences $X^j$ strongly. A light-blue block, in contrast, refers to a weak connection, and the lightest blue indicates no direct causal relationship between the two variables. For instance, 0.82 in $\mathbf{ME_1(GC)}$ locates in position $(X_1, X_2)$, representing $X_1$ causes $X_2$ and the causal strength is 0.82. Moreover, the dark blue block shows a strong causal relationship. The comparison graphs reveal that the outcomes from diverse causal detectors are different. The PCMCI+ discovers all causal links correctly. The GC performs well in processing linear systems but the indirect adjacency $X^1 \to X^3$ is also identified. The NTE and CCM do not perform well in linear systems. Therefore, the following steps of the rule ensemble and optimization are significant for improving the stability and trustworthiness of causal inference.

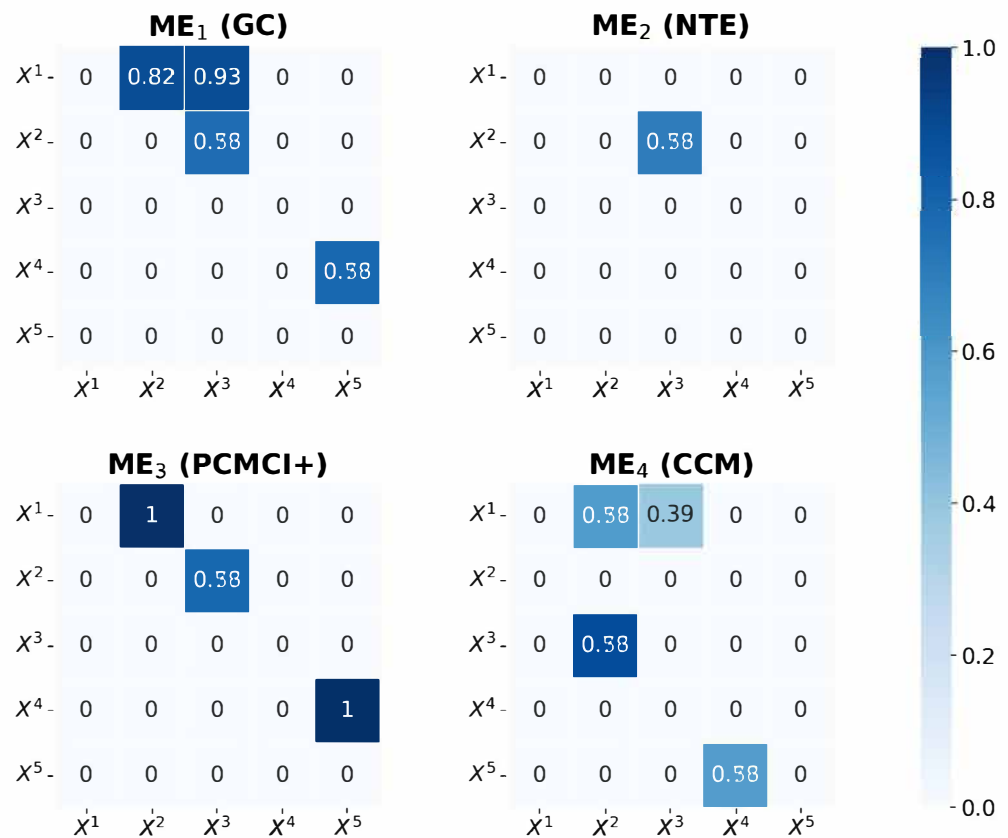

**Figure 6.** GMM-ensemble results of dataset 1. The four matrices represent four casual results of distinct base learners after GMM-processing. The element in position $(X^i, X^j)$ demonstrates that $X^i$ influences $X^j$, and the digital number refer to the causal strength, where 0 denotes none causality and 1 denotes strongest causality (e.g., 0.82 in $\mathbf{ME_1(GC)}$ represents that $X^1$ causes $X^2$ and causal strength is 0.82). The color legend represents the correspondence between causal strength and block color.

Figure 7 shows the final result of the proposed causal ensemble model and the causality ground truth for comparison. It indicates that all causal links in dataset 1 are derived and no false positives are detected. Figure 8 is the causal graph of dataset 1 based on the left plot in Figure 7. The arrow denotes the direction of causality, and the shades of blue represent the causal strength. The dark-blue arrow indicates that the causality in this direction is strong, whereas the light-blue arrow reveals a weak causal relationship.

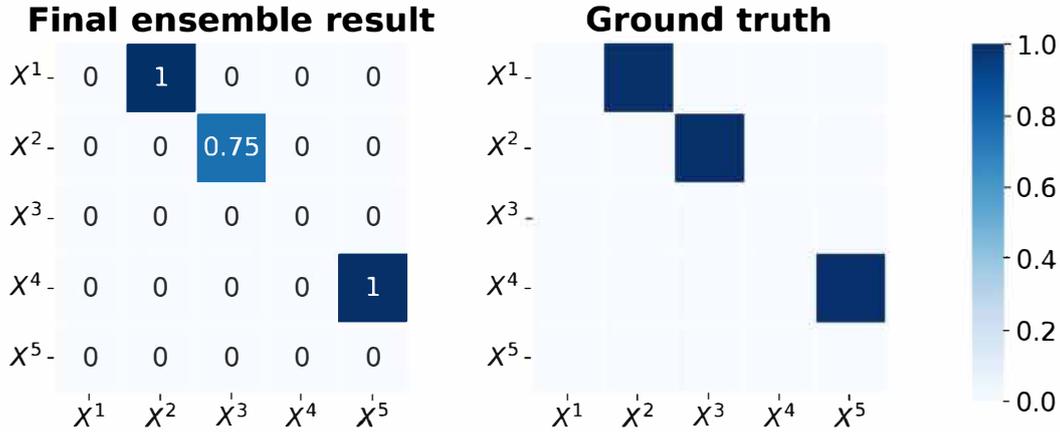

**Figure 7.** Final causal strength matrix and the ground truth of dataset 1. The left sub-figure is the final quantitative causal results, and its structure is same as those in Figure 6. The right sub-figure is the ground truth based on Equation (27), which denotes the setting causal relationship qualitatively. The dark-blue block represents causality whereas the light-blue block refers to non-causality. Quantitative causal strengths are only generated by causal inference algorithms, and are not defined within the dataset itself. Therefore, there is no numerical representation in the right sub-figure.

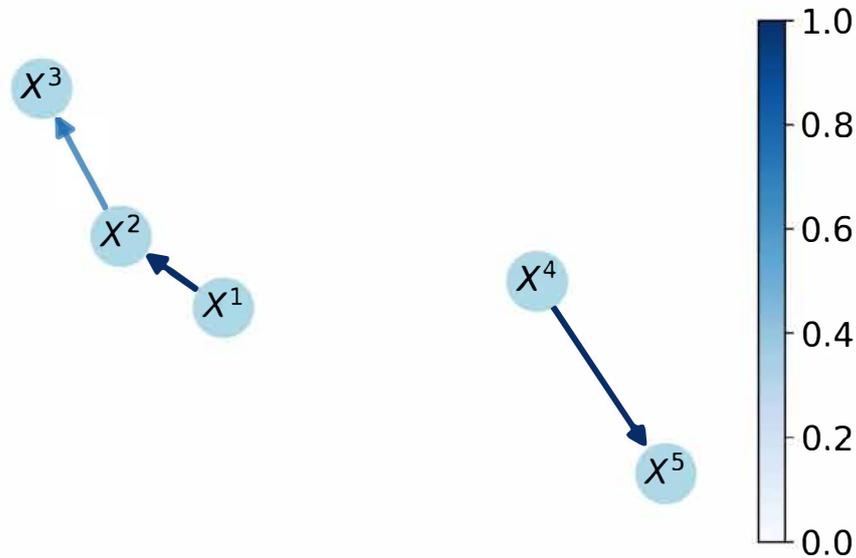

**Figure 8.** Causal graph of dataset 1. $X^1, \ldots, X^5$ in the circle represent distinct time series. The arrow denotes the direction of causality, where the shades of blue represent the causal strength and the correspondence is denoted in the color legend.

The evaluation index, Credibility Score ($CS$) is 0.8, which means that this result can be 80% trusted and that the credibility is strong.

Now, looking at dataset 2, Figure 9 illustrates the results of the GMM ensemble phase. The PCMCI+ and CCM both detect three out of four causal links, but their conclusions are different. The NTE discovers one correct causality pair and misses three. The GC derives all cause–effect pairs, as well as three false positives. It is not considered a convincing detection because of the false positives. However, after the rule ensemble and optimization, as Figures 10 and 11 illustrates, the final outcome is satisfying. Specifically, it discovers all correct causal connections without any false positive values. In the evaluation index, $CS$ is 0.8 which signals that this result can be 80% trusted (strong credibility).

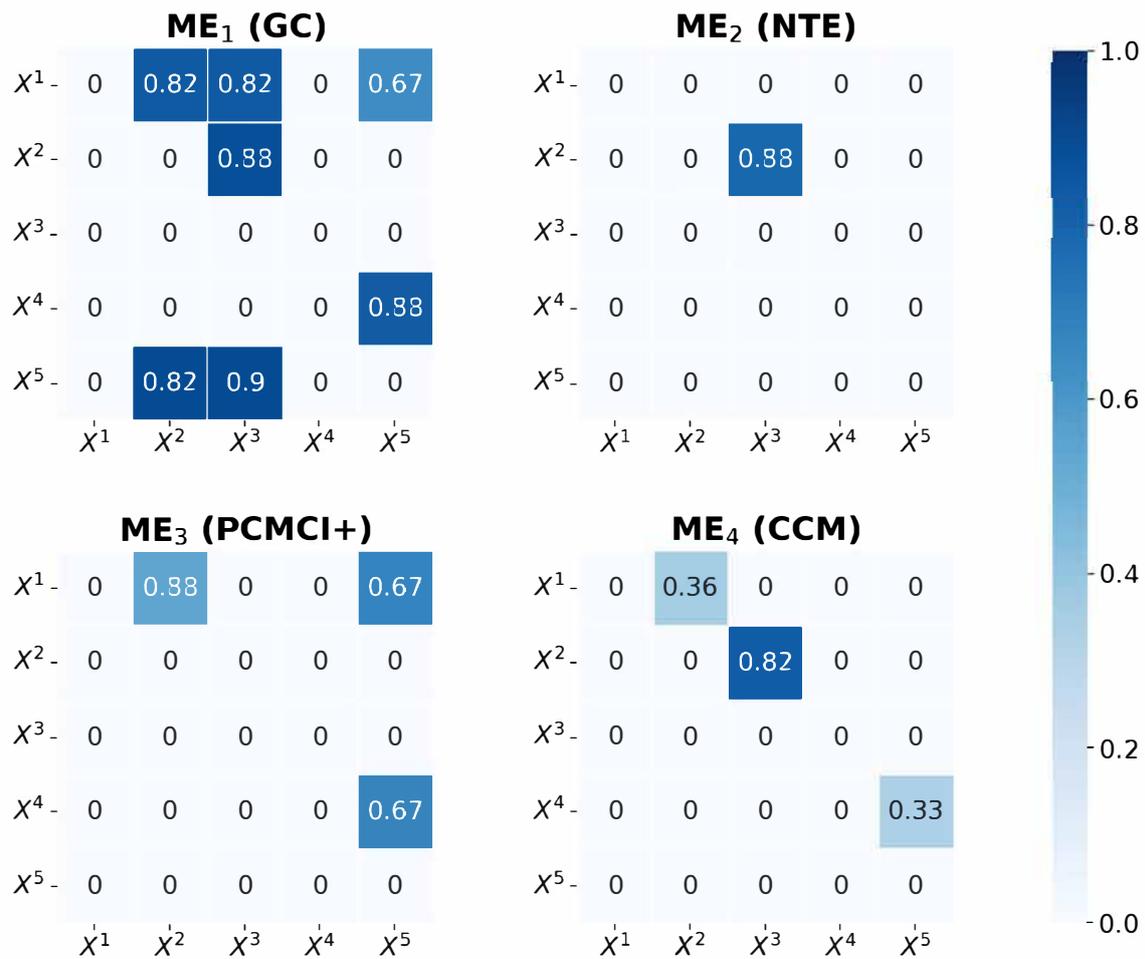

**Figure 9.** GMM-ensemble results of dataset 2. The four matrices represent four casual results of distinct base learners after GMM-processing. The element in position $(X^i, X^j)$ represents that $X^i$ influences $X^j$, and the digital number refer to the causal strength where 0 denotes none causality and 1 denotes strongest causality (e.g., 0.83 in $\mathbf{ME_1(GC)}$ represents that $X^1$ causes $X^2$ and causal strength is 0.83). The color legend represents the correspondence between causal strength and block color.

|     | $X^1$ | $X^2$ | $X^3$ | $X^4$ | $X^5$ |
|-----|-------|-------|-------|-------|-------|
| $X^1$ | 0 | 0.59 | 0 | 0 | 0.67 |
| $X^2$ | 0 | 0 | 0.59 | 0 | 0 |
| $X^3$ | 0 | 0 | 0 | 0 | 0 |
| $X^4$ | 0 | 0 | 0 | 0 | 0.59 |
| $X^5$ | 0 | 0 | 0 | 0 | 0 |

**Final ensemble result** / **Ground truth**

**Figure 10.** Final causal strength matrix and the ground truth of dataset 2. The left sub-figure is the final quantitative causal results, and its structure is same as those in Figure 9. The right sub-figure is the ground truth based on Equation (28), which denotes the setting causal relationship qualitatively. The dark-blue block represents causality whereas the light-blue block refers to non-causality. Quantitative causal strengths are only generated by causal inference algorithms, and are not defined within the dataset itself. Therefore, there is no numerical representation in the right sub-figure.

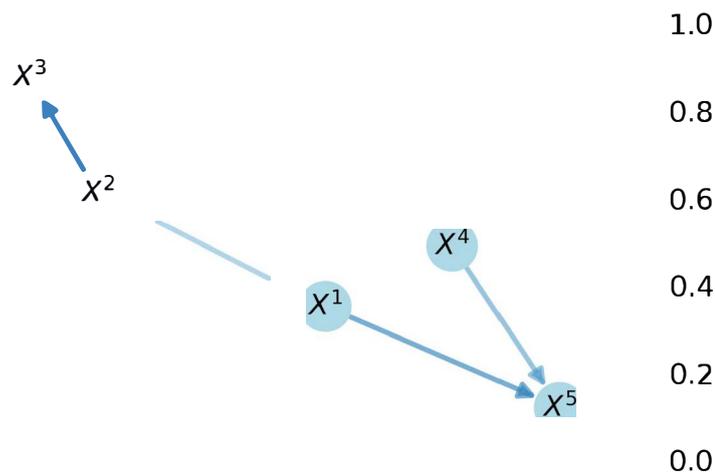

**Figure 11.** Causal graph of dataset 2. $X^1, \ldots, X^5$ in the circle represent distinct time series. The arrow denotes the direction of causality, where the shades of blue represent the causal strength and the correspondence is denoted in the color legend.

Dataset 3 is an exceedingly complex, non-linear system. Figure 12 displays the outcomes of the GMM ensemble phase. with the increasing complexity of the datasets, the GC cannot derive a credible conclusion although it discovers all the correct cause–effect pairs. This is because it also detects many false positives. The NTE detects nearly half the correct causal links with only one false positive result. The CCM discovers more than half correct causal effects but identifies 9 false positives as well. The PCMCI+ performs better than the other approaches in such a complex dataset, finding 10 out of the 12 causal links without any false positives. After the rule ensemble and optimization, as Figures 13 and 14 illustrates, the final result is more convincing than any of the four base learners. Indeed, it discovers all correct causal connections with only two false positives. In the evaluation index, *CS* is 0.814 which indicates that this causality result can be 81.4% trusted (strong credibility).

*5.4. Comparison and Evaluation*

This section quantitatively evaluates the performance of the proposed causal ensemble model. To perform that evaluation, a confusion matrix is defined based on the causality ground truth. The confusion matrix is originally used to visualize the performance of supervised learning algorithm for classification tasks. In causal inference, the synthetic dataset processing involves defining the causality ground truth as the actual condition. In such instances, the causal effect denotes the positive value (P) and non-causality represents the negative value (N). After this step, the four outcomes, including true positives (TP), false negatives (FN), false positives (FP), and true negatives (TN), are generated and the indexes, including accuracy, precision, recall, and $F1$ score are computed.

In addition to comparing the model performance on diverse datasets, the two-phase ensemble result is compared with the GMM-ensemble results of the four base causal detectors. This is performed to evaluate whether the two-phase ensemble functions better than the single causal reasoning algorithm.

Tables 3–5 demonstrate the performance of the four base learners and the proposed two-phase causal ensemble model on datasets 1–3, respectively. The bold values represent that the relevant method performs best based on this specific evaluation index.

The proposed ensemble model ranks first in all indexes (accuracy, precision, recall, and $F1$ score) on all three datasets except on dataset 3, where it ranks second in precision. This indicates that the proposed algorithm performs better than the single causality discovery algorithm.

The complexity and volume of the three datasets differ but this ensemble model derives good causality conclusions on both linear and non-linear systems. This is true regardless of the number of variables and observations, which reveals its stability and generalization ability. The $CS$ indexes in all experiments are greater than 0.75, indicating that the causal detection can be strongly trusted. The evaluating results prove the stability and reliability of the proposed causal ensemble model.

Table 3. Performance comparison of the single causal inference model and the proposed causal ensemble model on dataset 1.

| Name | TP | FN | FP | TN | Accuracy | Precision | Recall | F1 Score | CS |
|---|---|---|---|---|---|---|---|---|---|
| GC | 3 | 0 | 1 | 16 | 0.95 | 0.75 | **1.0** | 0.86 | / |
| NTE | 1 | 2 | 0 | 17 | 0.9 | **1.0** | 0.33 | 0.5 | / |
| PCMCI+ | 3 | 0 | 0 | 17 | **1.0** | **1.0** | **1.0** | **1.0** | / |
| CCM | 1 | 2 | 3 | 14 | 0.75 | 0.25 | 0.33 | 0.28 | / |
| Ensemble model | 3 | 0 | 0 | 17 | **1.0** | **1.0** | **1.0** | **1.0** | 0.8 |

Table 4. Performance comparison of the single causal inference model and the proposed causal ensemble model on dataset 2.

| Name | TP | FN | FP | TN | Accuracy | Precision | Recall | F1 Score | CS |
|---|---|---|---|---|---|---|---|---|---|
| GC | 4 | 0 | 3 | 13 | 0.85 | 0.57 | **1.0** | 0.73 | / |
| NTE | 1 | 3 | 0 | 16 | 0.85 | **1.0** | 0.25 | 0.4 | / |
| PCMCI+ | 3 | 1 | 0 | 16 | 0.95 | **1.0** | 0.75 | 0.86 | / |
| CCM | 3 | 1 | 0 | 16 | 0.95 | **1.0** | 0.75 | 0.86 | / |
| Ensemble model | 4 | 0 | 0 | 16 | **1.0** | **1.0** | **1.0** | **1.0** | 0.8 |

Table 5. Performance comparison of the single causal inference model and the proposed causal ensemble model on dataset 3.

| Name | TP | FN | FP | TN | Accuracy | Precision | Recall | F1 Score | CS |
|---|---|---|---|---|---|---|---|---|---|
| GC | 12 | 0 | 26 | 94 | 0.8 | 0.32 | **1.0** | 0.48 | / |
| NTE | 5 | 7 | 1 | 119 | 0.94 | 0.83 | 0.42 | 0.56 | / |
| PCMCI+ | 10 | 2 | 0 | 120 | **0.98** | **1.0** | 0.83 | 0.91 | / |
| CCM | 7 | 5 | 9 | 111 | 0.89 | 0.44 | 0.58 | 0.5 | / |
| Ensemble model | 12 | 0 | 2 | 118 | **0.98** | 0.86 | **1.0** | **0.93** | 0.81 |

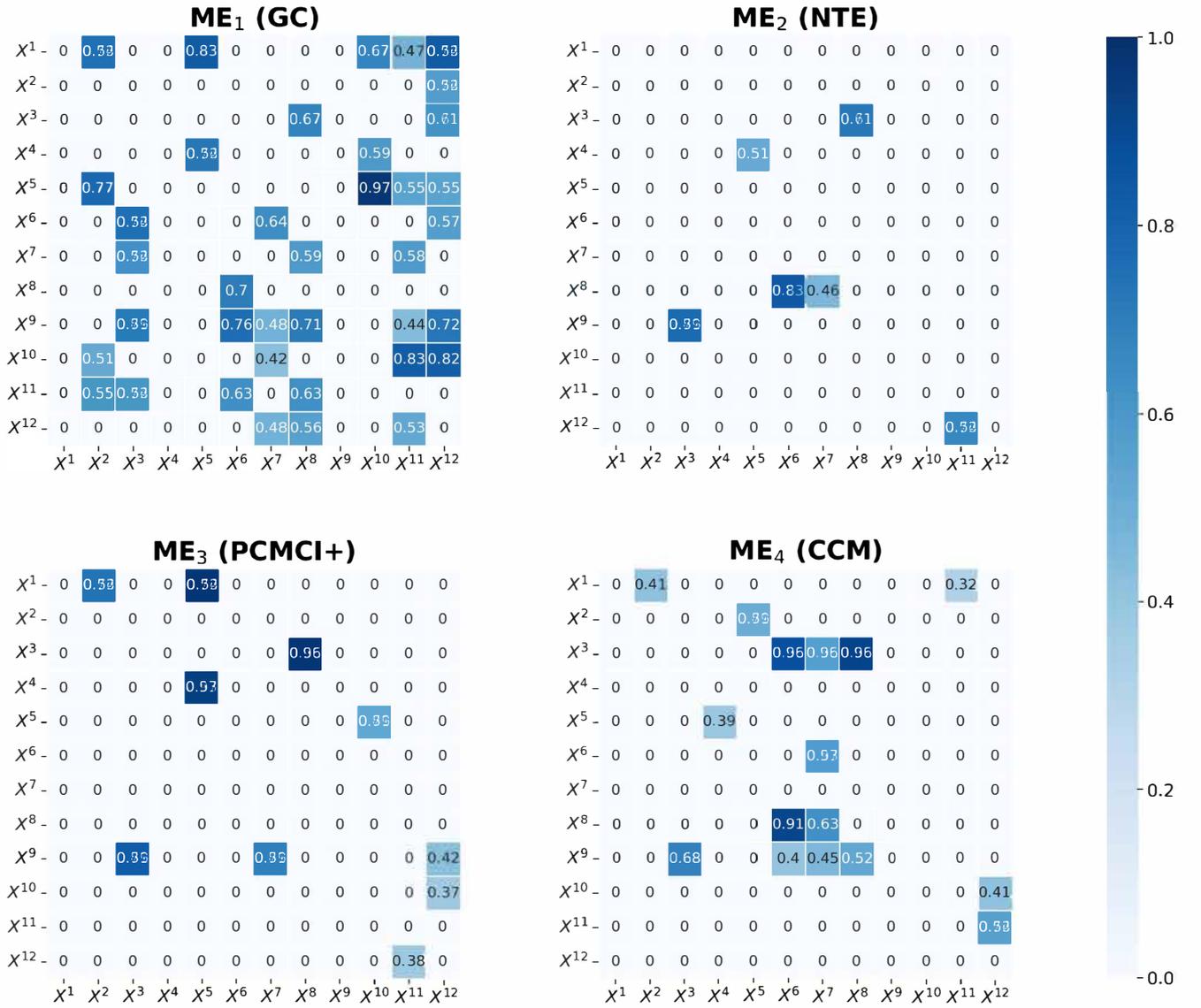

**Figure 12.** GMM-ensemble results of dataset 3. The four matrices represent four casual results of distinct base learners after GMM-processing. The element in position $(X^i, X^j)$ represents that $X^i$ influences $X^j$, and the digital number refer to the causal strength where 0 denotes none causality and 1 denotes strongest causality (e.g., 0.74 in $\mathbf{ME_1(GC)}$ represents that $X^1$ causes $X^2$ and causal strength is 0.74). The color legend represents the correspondence between causal strength and block color.

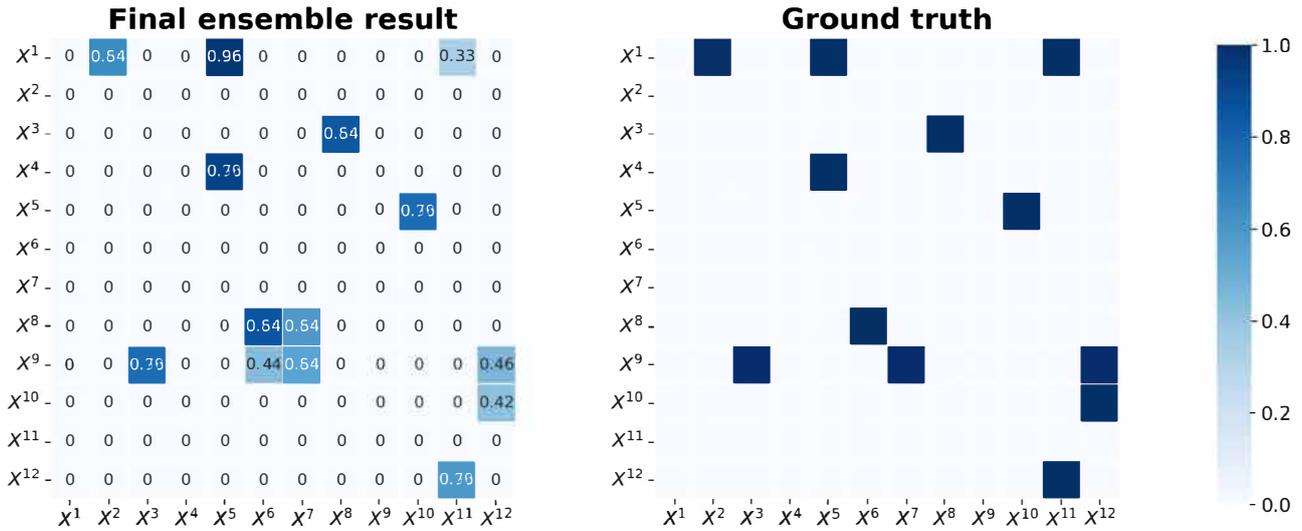

**Figure 13.** Final causal strength matrix and the ground truth of dataset 3. The left sub-figure is the final quantitative causal results, and its structure is same as those in Figure 12. The right sub-figure is the ground truth based on Equation (29), which denotes the setting causal relationship qualitatively. The dark-blue block represents causality whereas the light blue block refers to non-causality. Quantitative causal strengths are only generated by causal inference algorithms, and are not defined within the dataset itself. Therefore, there is no numerical representation in the right sub-figure.

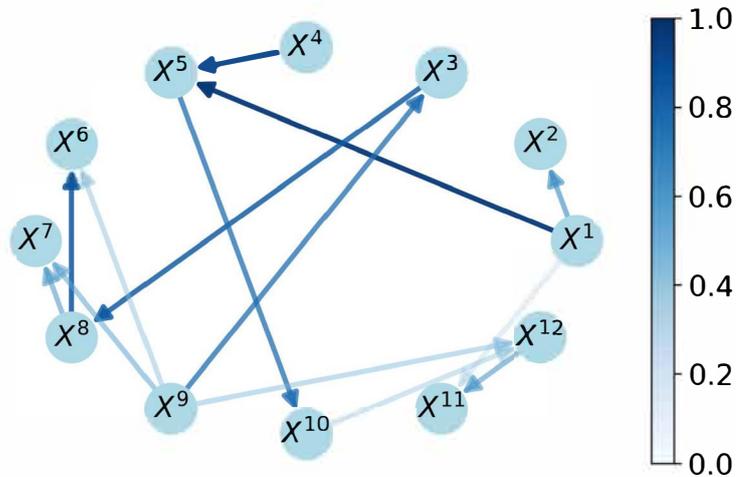

**Figure 14.** Causal graph of dataset 3. $X^1, \ldots, X^{12}$ in the circle represent distinct time series. The arrow denotes the direction of causality, where the shades of blue represent the causal strength and the correspondence is denoted in the color legend.

*5.5. Limitations and Future Work*

The experiments have produced dependable and persuasive results for our causal ensemble model, but there are still some aspects that can be further improved. Firstly, the present model fixes the time lag, which may not be optimal for the system. A method that considers dynamic time-lag selection could be developed in future work. Secondly, the rules in the second ensemble phase could be further enhanced by incorporating more comprehensive conditions. Lastly, while the evaluation index $CS$ is suitable for assessing this specific model, a more universal and precise statistical index could be developed for broader evaluation purposes.

## 6. Conclusions

We presented an ensemble approach for causal inference consisting of two phases. In the first phase, parallel computation of individual causal inference methods is enabled by partitioning the given datasets into multiple overlapping subsets. This enables a speed-up while limiting the information loss resulting from the partitioning. The results computed on individual partitions are then combined by applying GMM clustering. This process is repeated for every one of the four base learners in the ensemble, resulting in four different intermediate results.

In the second phase, these intermediate results are then integrated by comprehensive evaluation and comparison based on three rules. This step takes advantage of the strength of the base learners and avoids their weaknesses as much as possible.

The proposed ensemble framework is a general and flexible structure, meaning that the base learners can be freely exchanged. In our experiments, the causal ensemble model improves the reliability of time series causality discovery of the four employed causal inference models (GC, NTE, PCMCI+, and CCM). The experiments yield good accuracy and well-controlled false positives.

Next to the ensemble approach itself, we make another contribution by developing an evaluation approach for ensemble causal inference algorithms, which we call the credibility score ($CS$). It is designed to assess the credibility of causal ensemble models by comparing the final ensemble result to the intermediate results from distinct base learners. Our proposed ensemble model achieves high credibility based on the evaluation index $CS$ presented in Section 4.

The performance of our ensemble approach may be further improved by extending it to combine causal links of different time lags. Further, the rule ensemble phase can be enhanced by introducing additional and more comprehensive rules. Thirdly, a more general and accurate statistical index can be developed to evaluate the credibility of causal inference.

**Author Contributions:** Conceptualization, Z.M., M.K. and C.E.; methodology, Z.M., M.K. and C.E.; software, Z.M.; validation, Z.M., M.K., D.B. and C.E.; formal analysis, Z.M., M.K., D.B. and C.E.; investigation, Z.M., M.K., D.B. and C.E.; resources, Z.M., M.K., D.B., C.E., D.L. and R.H.S.; data curation, Z.M., M.K., D.B. and C.E.; writing—original draft preparation, Z.M.; writing—review and editing, Z.M., M.K., D.B., C.E., D.L. and R.H.S.; visualization, Z.M., M.K., D.B. and C.E.; supervision, D.L. and R.H.S.; project administration, R.H.S.; funding acquisition, R.H.S. All authors have read and agreed to the published version of the manuscript.

**Funding:** Funded by the Deutsche Forschungsgemeinschaft (DFG, German Research Foundation) under Germany's Excellence Strategy—EXC-2023 Internet of Production—390621612.

**Data Availability Statement:** The data and the code for data analysis that support the findings of this study are openly available in the repository—'two-phase-causal-ensemble-model' [45] at https://github.com/zhipengmichaelma/two-phase-causal-ensemble-model. (accessed on 28 March 2023).

**Conflicts of Interest:** The authors declare no conflicts of interest.